\PassOptionsToPackage{table}{xcolor}
\documentclass[sigconf=true, nonacm=true, review=false, anonymous = false,]{acmart}
\usepackage{booktabs} 

\copyrightyear{2021}
\acmYear{2021}
\setcopyright{none}
\acmConference[GECCO '21]{Genetic and Evolutionary Computation Conference}{July 10--14, 2021}{Lille, France}

\usepackage{amsmath}
\usepackage{xspace}
\usepackage{xcolor}
\usepackage{dsfont}
\usepackage{subcaption}
\usepackage{rotating}
\usepackage{makecell}

\usepackage{amsthm}
\usepackage{amsmath}
\renewcommand{\epsilon}{\varepsilon}

\newcommand{\fbest}{\ensuremath{f_{\text{best}}}}

\newcommand{\mean}[1]{\ensuremath{\operatorname{ave}(#1)}}
\newcommand{\norm}[1]{\left\lVert#1\right\rVert}

\setcopyright{rightsretained}
\acmYear{2021}
\copyrightyear{2021}
\acmDOI{10.1145/nnnnnnn.nnnnnnn} 
\acmISBN{978-x-xxxx-xxxx-x/YY/MM} 

\acmConference[GECCO '21]{The Genetic and Evolutionary Computation Conference 2021}{July 10--14, 2021}{Lille, France}


\begin{document}
\title{Explorative Data Analysis of Time Series based Algorithm Features of CMA-ES Variants}

\author{Jacob de Nobel}
\affiliation{%
  \institution{Leiden University}
  \city{Leiden}
  \country{Netherlands}
}
\author{Hao Wang}
\affiliation{%
  \institution{Leiden University}
  \city{Leiden}
  \country{Netherlands}
}
\author{Thomas B\"ack}
\affiliation{%
  \institution{Leiden University}
  \city{Leiden}
  \country{Netherlands}
}

\renewcommand{\shortauthors}{J. de Nobel et al.}

\begin{abstract}
In this study, we analyze behaviours of the well-known CMA-ES by extracting the time-series features on its dynamic strategy parameters. An extensive experiment was conducted on twelve CMA-ES variants and 24 test problems taken from the BBOB (Black-Box Optimization Bench-marking) testbed, where we used two different cutoff times to stop those variants. We utilized the \texttt{tsfresh} package for extracting the features and performed the feature selection procedure using the \texttt{Boruta} algorithm, resulting in 32 features to distinguish either CMA-ES variants or the problems. After measuring the number of predefined targets reached by those variants, we contrive to predict those measured values on each test problem using the feature. From our analysis, we saw that the features can classify the CMA-ES variants, or the function groups decently, and show a potential for predicting the performance of those variants. We conducted a hierarchical clustering analysis on the test problems and noticed a drastic change in the clustering outcome when comparing the longer cutoff time to the shorter one, indicating a huge change in search behaviour of the algorithm. In general, we found that with longer time series, the predictive power of the time series features increase.
\end{abstract}

\maketitle

\section{Introduction} \label{sec:intro}
In the field of continuous black-box optimization, a large variety of optimization algorithms are proposed in the past~\cite{opt1,opt2} and some prominent ones, e.g., the Covariance Matrix Adaptation Evolution Strategy (CMA-ES)~\cite{cmaes} have been successfully applied to solve real-world problems. More recently, many interesting scientific efforts have been devoted to these algorithms to boost up their performance even further. Particularly, for the CMA-ES, many variants have been devised based on the standard algorithm~\cite{cmaes}, e.g., the so-called BIPOP-CMA-ES~\cite{bipop} (please see~\cite{vanRijn} for an overview). Also, it is known that their empirical performance is often hard to distinguish with a statistical meaning when aggregated over a large number of different test problems~\cite{vanRijn}.
This given, provides the need to tackle the well-known \textit{algorithm selection} problem~\cite{journals/ac/Rice76}, which targets to find a proper optimization algorithm among a portfolio on some problem class~\cite{mersmanBBOB}. As an effective approach into this direction, Explorative Landscape Analysis (ELA)~\cite{ela} claims that it is vital for the algorithm selection task to obtain knowledge on optimization problems. It contrives to acquire such knowledge on the properties of test problems by designing numerical features. Rather than focusing on the problem side, in this paper, we argue that it is equally important to construct a reliable and useful quantification of algorithm behaviour. Answering the question; \textit{how different optimization algorithms would behave on different optimization problems} since conceivably, an appropriate algorithm could be chosen more effective if we have sufficient knowledge from both the algorithm and problem side.

Here, the first step in this direction is taken by monitoring the strategy parameters of the CMA-ES variants during the optimization runs and extracting time series features from the dynamics of those parameters. Specifically, we adopted a wide set of pre-defined time series features as in the \texttt{tsfresh} library, e.g., the entropy or the absolute sum of changes of the series, and apply them on the strategy parameters of the CMA-ES, e.g., the step-size and the eigenvalues of the covariance matrix.
For instance, we could compute the so-called ``absolute sum of changes'' feature for the step-size, which measures how much the step-size oscillates over time. In this manner, time series analysis could provide insight into how the different CMA-ES variants behave over time by quantifying the differences in their features. 


Facilitated by this approach, we analyse the differences between several different CMA-ES variants on single or multiple test problems. We also delve into in the dynamics of a single variant's strategy parameters on different benchmark problems. This also provide us knowledge about the objective function a variant is optimizing, allowing us to classify objective functions in terms of algorithm dynamics. Furthermore, this could also open the possibility of quantifying problem properties, serving as as an alternative to the standard ELA methodology. We further extend our investigation by exploring the feasibility of predicting the performance of those variants using the its time series features. 

As for the experimental study, we took the well-know BBOB test problems from the COCO platform~\cite{nikolaus_hansen_2019_2594848} and conducted clustering analysis and predictive modeling for the time series features.

This paper is organized as follows; section~\ref{sec:background} introduces the necessary preliminaries for the proposed time series feature based approach, described in section~\ref{sec:approach}, including a hierarchical clustering experiment. Section~\ref{sec:classification} includes a supervised learning experiment with the computed time series features. This is followed by a regression approach in section~\ref{sec:regression}, investigating the relation between the time series features and the benchmark performance of the CMA-ES. The paper is concluded in section~\ref{sec:conclusion}. 

\section{Preliminaries} \label{sec:background}
In this work, we focus on optimizing real-valued objective functions of the form $f\colon [-5,5]^d \rightarrow \mathbb{R}$ and take the well-known Black-Box Optimization-Benchmarking (BBOB) noiseless problem set~\cite{nikolaus_hansen_2019_2594848}. We take instance number $1$ and set the search dimension to $d = 5$. 


\paragraph{CMA-ES} In this work, we analyse the CMA-ES algorithm empirically, a nature inspired continuous black box optimizer. In the CMA-ES, candidate solutions are sampled according to a multivariate normal distribution $\mathcal{N}(\vec{m}, \sigma^2\mathbf{C})$. Through an iterative procedure, the CMA-ES adapts the parameters of this distribution gradually using selected solutions in the current iteration as well as historical information on the trajectory of its parameters. Specifically, the \emph{rank-$\mu$-update} (see~\cite{cmatut} for a detailed explanation) yields the maximum likelihood estimate of $\mathbf{C}$ based on the selected solutions while the \emph{rank-one-update} utilizes the so-called cumulative evolution path $\vec{p}_c$, which summarizes the trajectory of the algorithm and preserves the sign of consecutive steps therein. Similarly, the step size $\sigma$ is controlled dynamically according to the length of the conjugate evolution path $\vec{p}_\sigma$, which accumulates consecutive steps transformed by the square root of the inverse covariance matrix $\mathbf{C}^{-\frac{1}{2}}$.  Hereinafter, we shall denote by $\mathbf{D}$ the diagonal matrix holding the eigenvalues of $\mathbf{C}$.

\paragraph{The Modular CMA-ES} 
The CMA-ES is a state of the art continuous black-box optimization algorithm, for which over the years several variants have been proposed in the literature~\cite{jastrebski2006improving, auger2011mirrored, orthogonal, msr, tpa, qrandom, thresholdConv}. We build upon the Modular EA (ModEA) framework~\cite{VanEvolving}, which contains eleven structural variants of the CMA-ES algorithm and implements those as configurable modules\footnote{nine of which takes binary options and the remaining two are ternary. Please see~\cite{VanEvolving} for the detail}. Taken as algorithmic building blocks, these modules can be used to create 4\,608 unique CMA-ES variants, which allows the user to easily construct custom variants of the CMA-ES algorithm. In ~\cite{VanEvolving} it was shown that this includes variants that outperform any classical CMA-ES variants from the literature. 
 
In this study, an extended version of ModEA, specifically designed for creating CMA-ES variants is used\footnote{The extended framework can be found here:Anonymized for review purposes.}. This Modular CMA-ES includes a number of extra options, specific to the CMA-ES. This includes an extra module for performing step size adaptation, the so-called Median Success Rule (MSR)~\cite{msr} step size adaptation rule. 
    
In this study, twelve CMA-ES variants are compared. The variants consist mostly of the all single-module\footnote{CMA-ES configurations with only one active module.} variants available within the Modular CMA-ES, with a few exceptions. Most notably, these are modules that cause every generation to have a different number of fitness function evaluations, such as modules that employ a restart strategy (e.g. IPOP) or the module for sequential selection. These modules are left out because strategy parameters are updated only once every generation and if a variant is allowed to use more (or less) fitness function evaluations per generation than other variants, this would create a bias in the observations of the strategy parameters, which is why such variants are left out. The list includes:

\begin{enumerate}
    \item \emph{Standard}: The standard CMA-ES, as defined in ~\cite{cmaes}. 
    \item \emph{Active}: CMA-ES with active covariance matrix update~\cite{jastrebski2006improving} penalizes bad candidate solutions with negative weights in the covariance matrix update.
    \item \emph{Mirrored} sampling~\cite{mirrored1}:  For each new sampled point, a mirrored pair is produced.  
    \item \emph{Mirrored Pairwise} sampling~\cite{auger2011mirrored, orthogonal}: Mirrored sampling, however, only the best of the two mirrored points is selected. 
    \item \emph{Orthogonal} sampling: The sampled candidate solutions are orthonormalized using a Gram-Schmidt procedure~\cite{orthogonal}.  
    \item \emph{Elitist}: $(\mu + \lambda)$ - selection. 
    \item \emph{Equal Weights}: Recombination weights of $\frac{1}{\lambda}$  
    \item \emph{MSR}: Median success rule step size adaptation~\cite{msr}. 
    \item \emph{TPA}: Two-point step size adaptation~\cite{tpa}. 
    \item \emph{Halton} sampling: Quasi-random Halton sampling~\cite{qrandom}.
    \item \emph{Sobol} sampling: Quasi-random Sobol sampling~\cite{qrandom}.
    \item \emph{Threshold Convergence}: Threshold Convergence of mutations~\cite{thresholdConv}
\end{enumerate}

\paragraph{CMA-ES' dynamic parameters} Many internal parameters of the CMA-ES algorithms are adapted throughout the optimization process, e.g., the center of mass and the covariance matrix. Among all such parameters, we consider, in this work, the following ones for our analysis: 
\begin{itemize}
    \item $\sigma \in \mathbb{R}_{\geq 0}$: The step size.
    \item $\fbest \in \mathbb{R}$: The best-so-far objective value.
    \item $\vec{v} \coloneqq \left(\sqrt{\lambda_1},\ldots, \sqrt{\lambda_d}\right)^\top$:  The square root of the diagonal of $\mathbf{D}$ 
    \item $\vec{p}_c\in \mathbb{R}^d$: The evolution path.
    \item $\vec{p}_\sigma\in \mathbb{R}^d$: The conjugate evolution path. 
\end{itemize}
The covariance matrix $\mathbf{C}$ and the basis of its eigenvectors $\mathbf{B}$ are not considered in this experiment, as they would immensely increase the number of observations stored, while the information gained in storing these variables is small in comparison. Also, the center of mass $\vec{m}$ is also not taken into account, as we consider it less of a variable telling something about the state of the algorithm than about a location in the search space.

\paragraph{Time series features}
Here, we consider a multivariate real-valued time series with a fixed sampling frequency~\cite{chatfield2004timeseries}, which represents, in this context, the dynamic strategy parameters of the CMA-ES algorithm. In each iteration of CMA-ES, we compute one data point of this time series as follows: $\forall t\in[L]$,
\begin{equation}
\vec{x}_t = \left(\sigma, \fbest, ||\vec{v}||, ||\vec{p}_\sigma||, ||\vec{p}_c||,  \mean{\vec{v}}, \mean{\vec{p}_\sigma},\mean{\vec{p}_c}\right)^\top,
\end{equation}
where $||\vec{v}|| = \sqrt{\vec{v}^\top\vec{v}}$ and $L$ is the total number of iterations (running length) executed in our experiments and \mean{\cdot} stands for the average of the components of its argument. We compute the component-wise average and norm of each vector variable in order to store their information invariant to problem dimension. Hereafter, we shall denote this multivariate time series as $\mathbf{X} = [\vec{x}_1, \vec{x}_2, \ldots, \vec{x}_L]\in \mathbb{R}^{8 \times L}$.

\begin{figure*}[!ht]
    \centering
    \includegraphics[width=\linewidth]{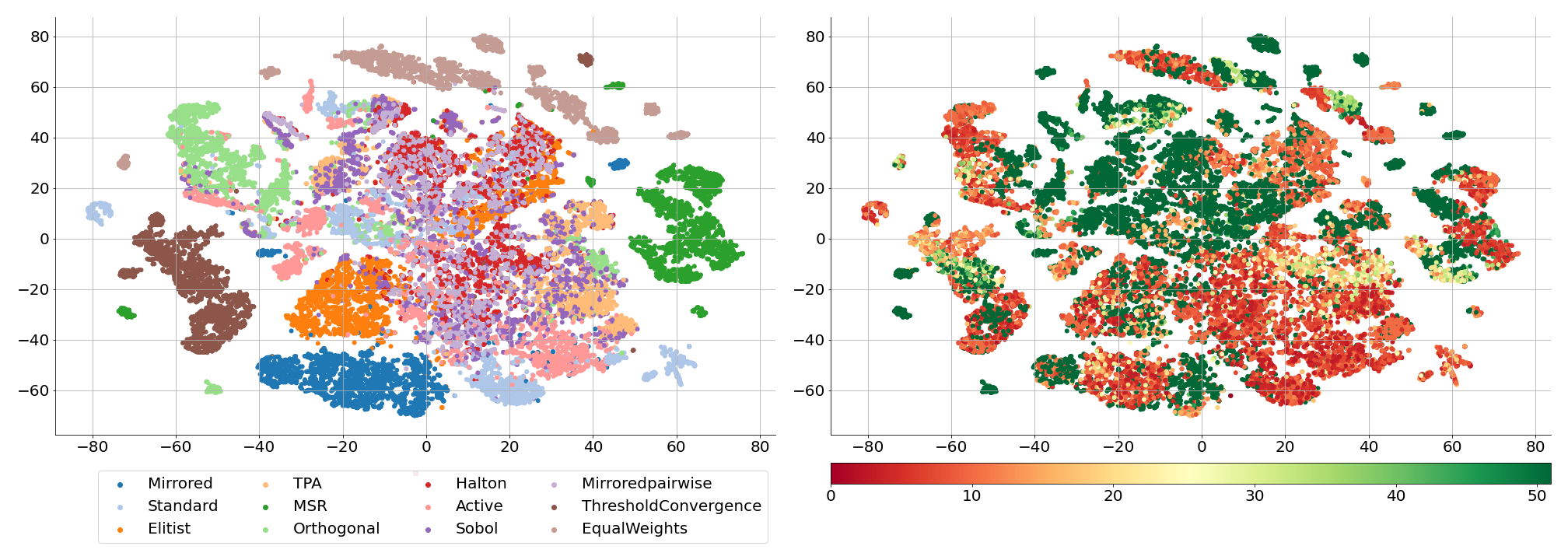}
    \caption{For the $L=500$ case, the two-dimensional embedding (obtained from t-SNE) of the selected time series features over all test problems are color-coded according to the variant (left) and the number of predefined targets $\{10^{2 - (x - 1) / 5} \colon x\in[0..51]\}$ being hit in each independent run (right).
    }
    \label{fig:distributionall}%
\end{figure*}
Given the time series data, we often probe into the characteristic thereof by means of extracting numerical features in the time domain, which give rise to meaningful interpretations and also embed the high-dimensional time series data into a lower-dimensional Euclidean space~\cite{tsfeatures}. Here, we adopted the widely used \texttt{tsfresh} package~\cite{tsfresh}, which implements a variety of feature functions for univariate time series, such as the auto-correlation and energy functions. Given a set of $m$ feature functions $\left\{\phi_i\right\}_{i=1}^m$ from \texttt{tsfresh} (where $\phi_i \colon \mathbb{R}^L \rightarrow \mathbb{R}$), we apply each feature function over each time series in $\mathbf{X}$ (i.e., each row vector of $\mathbf{X}$), yielding an overarching feature map, $\Phi \colon \mathbb{R}^{8\times L} \rightarrow \mathbb{R}^{8m}$:
\begin{align*}
\vspace{-10mm}
\mathbf{X} &\mapsto \left(\phi_1(\mathbf{X}_{1*}), \ldots,\phi_1(\mathbf{X}_{8*}),\ldots, \phi_m(\mathbf{X}_{1*}),\dots, \phi_m(\mathbf{X}_{8*})\right)^\top,
\end{align*}
where $\mathbf{X}_{i*}$ represents the $i$th row of $\mathbf{X}$. Hereafter, we shall denote a feature vector as $\Phi(\mathbf{X})$, or simply $\Phi$ if the input time series is clear from the context. The feature functions applied in our experiment can be found in table~\ref{table:features_explained}.
When applied in our case, such a feature extraction procedure, results in 6\,352 features (see section~\ref{subsec:data-generation} for detail), which ought to undergo a proper feature selection process (see section~\ref{subsec:feature-selection}) to reduce its dimensionality, hence facilitating the modelling tasks later.

\section{The proposed approach} \label{sec:approach}   
Given a benchmarking problem set $\mathcal{F} = \{f_1, f_2, \ldots\}$, a set of CMA-ES variants $\mathcal{A} = \{A_1, A_2, \ldots\}$, and a set of time series feature functions, we aim to distinguish those variants for each $f \in F$ by computing the aforementioned time series features of every variant on $f$ and building a classifier based on the features. Specifically, we repeat the feature computation $N$ times independently for each $A$ and $f$, since the time series features from a single run exhibits large variability. We shall denote by $\Phi(A, f, d, i)$ the time series features computed from the $i$th independent run of a variant $A$ on the $d$-dimensional function $f$. Taking this notation, we could express the data set for classifying (see section~\ref{sec:classification}) CMA-ES variants as
$\left\{(\Phi(A, f, d, i), A)\right\}_{i\in[1..N], A\in\mathcal{A},f\in\mathcal{F}},$
and similarly the one for discriminating test problems, $\left\{(\Phi(A, f, d, i), f)\right\}_{i\in[1..N], A\in\mathcal{A},f\in\mathcal{F}}$.
As previously mentioned, we consider $12$ CMA-ES variants for $\mathcal{A}$, $24$ BBOB problems for $\mathcal{F}$ with $N=100$ and $d=5$. For all the classification tasks presented in the sequel, we use a variant of the random forest model, called Extremely Randomized Trees Classifier (ERTC)~\cite{ExRandTrees}, which constructs each constituting decision tree without bootstrapping based on random splitting of internal nodes (implemented by the Python library scikit-learn~\cite{sklearn}). We took the default hyper-parameter setting of this algorithm since it is not the focus of this work to perform a hyperparameter tuning task and the default hyperparameter settings are empirically robust. We choose this model because it treats all features more equally compared to the random forest, since as an aggressive feature selection was applied before training this model (see Sec.~\ref{subsec:feature-selection}).

\subsection{Data generation}\label{subsec:data-generation}
To generate the time series data, we conducted $N=100$ independent runs for each pair of CMA-ES variant and test function in 5D, using the first instance (\texttt{iid = 1}) of each problem. 
In addition, for every run, we start all CMA-ES variants with the same initial mean, the step-size, and  covariance matrix (set to the identity matrix) to remove randomness incurred by the initialization, 
 making the time series yielded from different variants/problems more comparable. Each independent run was terminated with a fixed number of $500$ generations, from which two sets of features were extracted, one using only the first 100 generations (represented by $L=100$) of the data and the other using the complete length of 500 generations ($L=500$). In detail, we utilized the \texttt{tsfresh}~\cite{tsfresh} python library (version $0.16$) to extract the features of time series $\mathbf{X}$, where we took all $794$ default features\footnote{This huge set of features are obtained from the combination of $63$ default feature functions and several different parameterizations for each of them. Please see \url{https://tsfresh.readthedocs.io/en/latest/text/list_of_features.html} for the detail.},
resulting in $794\times 8 = 6352$ raw feature values after applied over each time series. 

Due to the fact that different feature values vary largely from one feature to the other and also from functions to functions, we re-scale the feature value to the unit interval for each time series on each function. Besides, we also recorded the number of predefined targets $N_s = \{10^{2 - (x - 1) / 5} \colon x\in[0..51]\}$ being hit in each independent run, for the purpose of predicting the performance of each CMA-ES variants based on its time series features.

\subsection{Feature selection} \label{subsec:feature-selection}
The vast number of raw features cast any learning task that takes them as input as a high-dimensional problem, and hence it is necessary to reduce its dimensionality by selecting more informative features. For this purpose, we picked the \texttt{Boruta}~\cite{boruta} algorithm\footnote{It is available in Python: \url{https://github.com/blue-yonder/tsfresh}.}, which works in conjunction with the Random Forests Classifier (RFC) and tries to capture all the important features in the data set w.r.t.~a target variable (here: the CMA-ES variant label). It creates so-called shadow features by randomly permuting the values for each feature, then trains a random forest\footnote{Essentially, any other feature importance calculation can also be applied here, e.g., from Gradient Boosting.} on the combination of the original features and the shadow features, and finally determines to retain a feature if its feature importance is significantly higher than its shadow counterpart. We took the default parameter settings of \texttt{Boruta} except that we set its maximal number of iterations to $150$ and took the ``auto''\footnote{Please see~\url{https://github.com/scikit-learn-contrib/boruta\_py} for more information on the Boruta package} option for the number of trees of the random forest model employed therein. Practically, we grouped the feature data by unique pairs of test problems and two running lengths $L\in\{100, 500\}$, which leads to $48$ groups, and performed the feature selection procedure on each group separately with \texttt{Boruta}, where a random forest model is trained to classify the algorithm variants on each test problem based on their selected time-series features. 

We only keep features that are selected by \texttt{Boruta} in all aforementioned $48$ groups of data. If multiple parameterizations of the same feature and time series combination are selected, only the combination that shows the highest feature importance is retained. The feature importance is retrieved from the random forest model underpinning the \texttt{Boruta} algorithm. This selection procedure managed to reduce the dimensionality drastically from $6\,352$ to 32, which is summarized in table~\ref{table:feature_sum} in terms of the number of selected ones for each raw time series, accompanied by the total and average feature importance (FI) values from the random forest model employed in \texttt{Boruta}.
From the table, we can observe that 1) none of the features originating from $f_{\text{best}}$ is selected; 2) features from $\norm{\vec{p}_\sigma}$ are selected the most frequently; 3) features based on $\sigma$ show the highest average feature importance. Table~\ref{table:features_explained} shows a specific overview of the selected feature functions and their parameter sets, as well as the time series they were applied over.  

Furthermore, we visualize the selected features of all test problems in Fig.~\ref{fig:distributionall}, which is obtained by embedding the features to a two-dimensional Euclidean space using the t-SNE~\cite{tsne} dimensionality reduction technique\footnote{We took the implementation from the sklearn~\cite{sklearn} library with the perplexity parameter set to 100.}. From the figure (left), we can see that many variants form distinct clusters in this embedded space, while some variants (e.g., Standard and Mirrored-Pairwise), however, are hardly distinguishable from each other. This is to be expected, since some variants are functionally more similar than other variants. 
\begin{table}[!ht]
 \small
\begin{tabular}{l|l|l|l}
 \toprule
   \textbf{Series}       & \textbf{\#selected features} & \textbf{Total FI} & \textbf{Average FI} \\ \hline
    $\norm{\vec{p}_\sigma}$ & \textbf{ 9}           & \textbf{1.1071}   & 0.1230              \\ \hline
    $\mean{\vec{p}_\sigma}$ & 1                     & 0.0398            & 0.0398              \\ \hline
    $\norm{\vec{p}_c}$      & 8                     & 0.6081            & 0.0760              \\ \hline
    $\mean{\vec{p}_c}$      & 0                     & -                 & -                \\ \hline
    $\norm{\vec{v}}$        & 5                     & 0.5442            & 0.1088              \\ \hline
    $\mean{\vec{v}}$        & 6                     & 0.8566            & 0.1428              \\ \hline
    $\sigma$                & 3                     & 0.4893            & \textbf{0.1631}     \\ \hline
    $f_{\text{best}}$              & 0                     & -                 & -                  \\ \hline 
    \bottomrule
\end{tabular}
\caption{Grouped by the time series, the number of selected features and their corresponding feature importance are listed with the highest value depicted in boldface.}
\label{table:feature_sum}
\end{table}
\begin{table}[!ht]
\hspace*{-6mm}
\footnotesize
\begin{tabular}{l|l|l}
    \hline
    Feature Function	&	Applied Over	&	Parameters\\ \hline
    absolute\_sum\_of\_changes	&	$\{\norm{\vec{p}_c}\}$	&	$\emptyset$ \\ \hline
    approximate\_entropy	&	$\{\norm{\vec{p}_\sigma}\}$	&	$\{(2, .1)\}$ \\ \hline
    autocorrelation	&	$\{\norm{\vec{p}_\sigma}\}$	&	$\{(1.)\}$ \\ \hline
    change\_quantiles	&	$\{\norm{\vec{p}_c}, \mean{\vec{p}_\sigma},\norm{\vec{p}_\sigma}, \sigma\}$ & 
    \makecell{$\{ (\texttt{mean}, \texttt{F}, .5, .6, .2)$,\\  $(\texttt{mean}, 1, .4, .6, 0)$, \\ $(\texttt{mean}, \texttt{F}, .4, .6, 0)$,\\ \ $(\texttt{mean}, \texttt{F}, .2, .6, 0) \}$} \\ \hline
    cid\_ce	&	$\{\norm{\vec{p}_\sigma}\}$	&	$\{ (\texttt{F}) \}$ \\ \hline
    energy\_ratio\_by\_chunks	&	$\{\norm{\vec{v}}, \mean{\vec{v}}\}$	&	$\{ (10, 0) \}$ \\ \hline
    fft\_aggregated	&	$\{\mean{\vec{v}}\}$	&	$\{ (\texttt{centroid}) \}$ \\ \hline
    fft\_coefficient	&	$\{\norm{\vec{p}_c}\}$	&	$\{(0, \texttt{abs}) \}$ \\ \hline
    index\_mass\_quantile	&	$\{\norm{\vec{v}}, \mean{\vec{v}}\}$	&	$\{ (.1)\}$ \\ \hline
    mean	&	$\{\norm{\vec{p}_c}\}$	& $\emptyset$ \\ \hline
    median	&	$\{\norm{\vec{v}}, \mean{\vec{v}}, \norm{\vec{p}_c}$, $\norm{\vec{p}_\sigma}\}$	&	$\emptyset$ \\ \hline
    minimum	&	$\{\norm{\vec{v}}, \mean{\vec{v}}\}$	&	$\emptyset$ \\ \hline
    number\_of\_crossing\_m	&	$\{\norm{\vec{p}_\sigma}\}$	&	$\{(1.)\}$ \\ \hline
    number\_peaks	&	$\{\sigma\}$	&	$\{(1.)\}$ \\ \hline
    partial\_autocorrelation	&	$\{\norm{\vec{p}_\sigma}\}$	&	$\{ (1.)\}$ \\ \hline
    quantile	&	$\{\norm{\vec{v}}, \mean{\vec{v}}, \norm{\vec{p}_c}$, $\norm{\vec{p}_\sigma}, \sigma\}$ & $\{(.1) \}$ \\ \hline
    range\_count	&	$\{\norm{\vec{p}_c}$, $\norm{\vec{p}_\sigma}\}$	&	$\{(1., -1.)\}$ \\ \hline
    sum\_values	&	$\{\norm{\vec{p}_c}\}$	&	$\emptyset$ \\ \hline
\end{tabular}
\caption{The selected time series features shown as the combination of the feature function, the time series on which the feature function is applied, and the parameterization therein. We express the parameterization as a set of tuples (empty for the parameterless case), where each tuple of parameter values is only applied on the time series appearing in the same order as with the tuple, e.g., for the ``change\_quantile'' function, $(\texttt{mean}, 0, .2, .6, 0) \}$ is only applied on $\sigma$. Please see~\url{https://tsfresh.readthedocs.io/en/latest/text/list\_of\_features.html} for the detail.}
\label{table:features_explained}
\end{table}

\subsection{Clustering Analysis}\label{distancemethod}

\begin{figure*}
\begin{subfigure}{0.45\textwidth}
   \includegraphics[width=\linewidth,trim=0 15mm 0 0,clip]{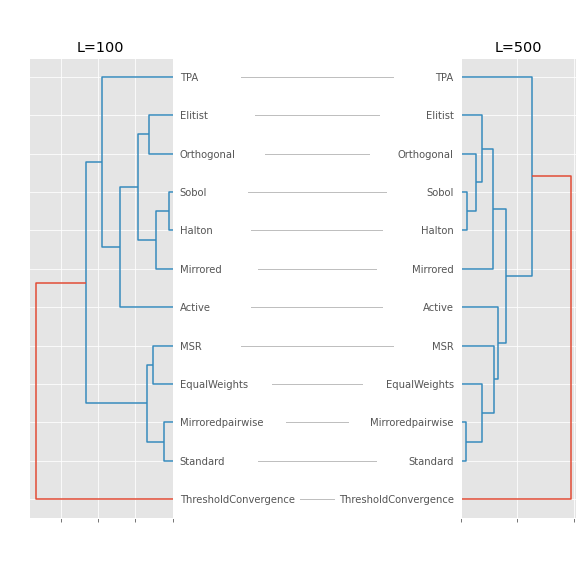}
 \end{subfigure}
 \hfill
 \begin{subfigure}{0.45\textwidth}
    \includegraphics[width=\linewidth, trim= 0 15mm 0 0,clip]{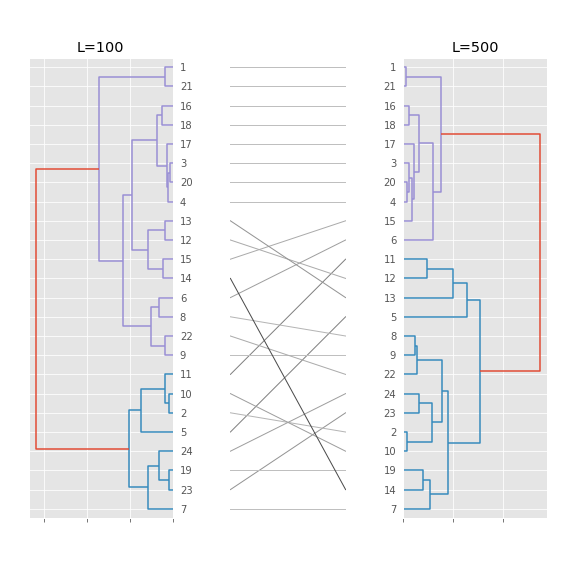}
 \end{subfigure}
 \caption{The hierarchical clustering results obtained on the pairwise distance of CMA-ES variants (left) and of test problems (right), in which we also contrast the $L=100$ case with $L=500$ to demonstrate the extent to which the clustering result would be affected by a prolonged running length. Also, we take the distance between the cluster to determine the color scale for the line connecting them. The dendrograms were produced for the mean vector of each variant/test function, using Euclidean distance and the Ward~\cite{ward} hierarchical clustering.}
\label{fig:tanglegramvariants}%
 
\end{figure*}

\begin{figure}%
    \centering
    \includegraphics[width=.7\linewidth, trim={40cm 0 20cm 0},clip]{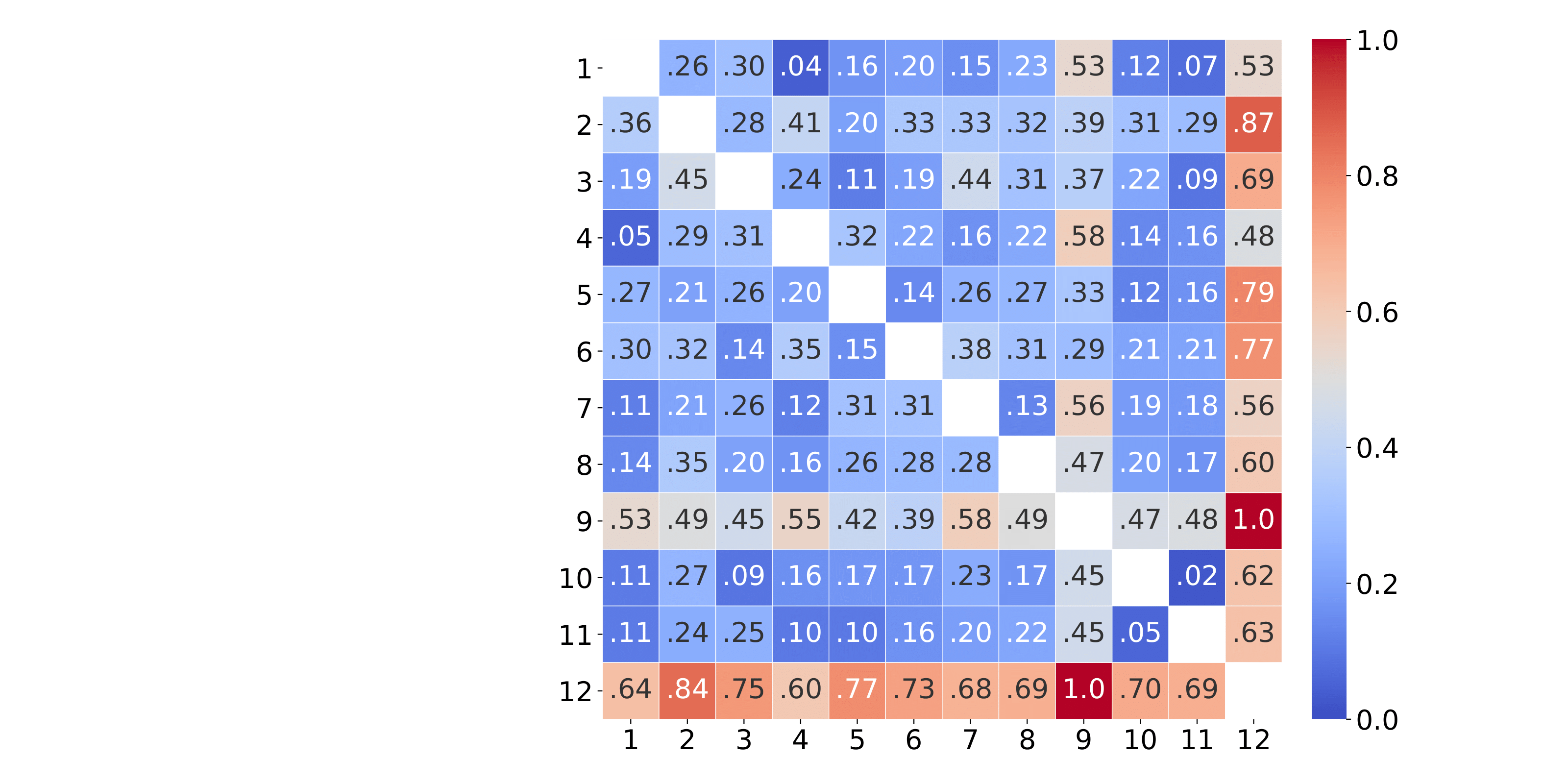}
    \caption{Pairwise distance matrix for all CMA-ES variants. The numeric index used here refers to the number specified in the list in section~\ref{sec:background}. Euclidean distance is computed and scaled though min-max normalization. The lower triangle of the figure shows the pairwise distances for $L=100$, the upper triangle shows $L=500$.} 
    \label{fig:distVarall}%
\end{figure}

Here, we propose to measure the similarity between CMA-ES variants by taking the mean of corresponding feature vectors over all test problems and runs, and then computing the mean Euclidean distance among the resultant. We render the distance matrix as a heatmap in Fig.~\ref{fig:distVarall} for both $L=100$ (the lower triangle) and $L=500$ (the upper triangle), which we observe that TPA and Threshold Convergence are the most dissimilar pair of variants, which are also quite distant from the other variants. For $L=100$, the most similar ones are the Sobol and Halton variants. Even though they still show a very high similarity for $L=500$, the most similar pair is the Standard CMA-ES and Mirrored Sampling with pairwise selection. In addition, we ran a more fine-grained hierarchical clustering on the distance matrix (shown in Fig.~\ref{fig:tanglegramvariants}, left).     
The similarities between the variants invoked the following consideration: \textit{variants which are seemingly different in terms of algorithmic structure, can exhibit similar behaviour}. This suggests that it can be helpful to probe into the dynamics of newly designed variants using time-series features, if such a variant does not perform as expected.

\begin{figure}[!ht]
\includegraphics[width=.5\textwidth,  trim={3.5cm 0 21cm 0},clip]{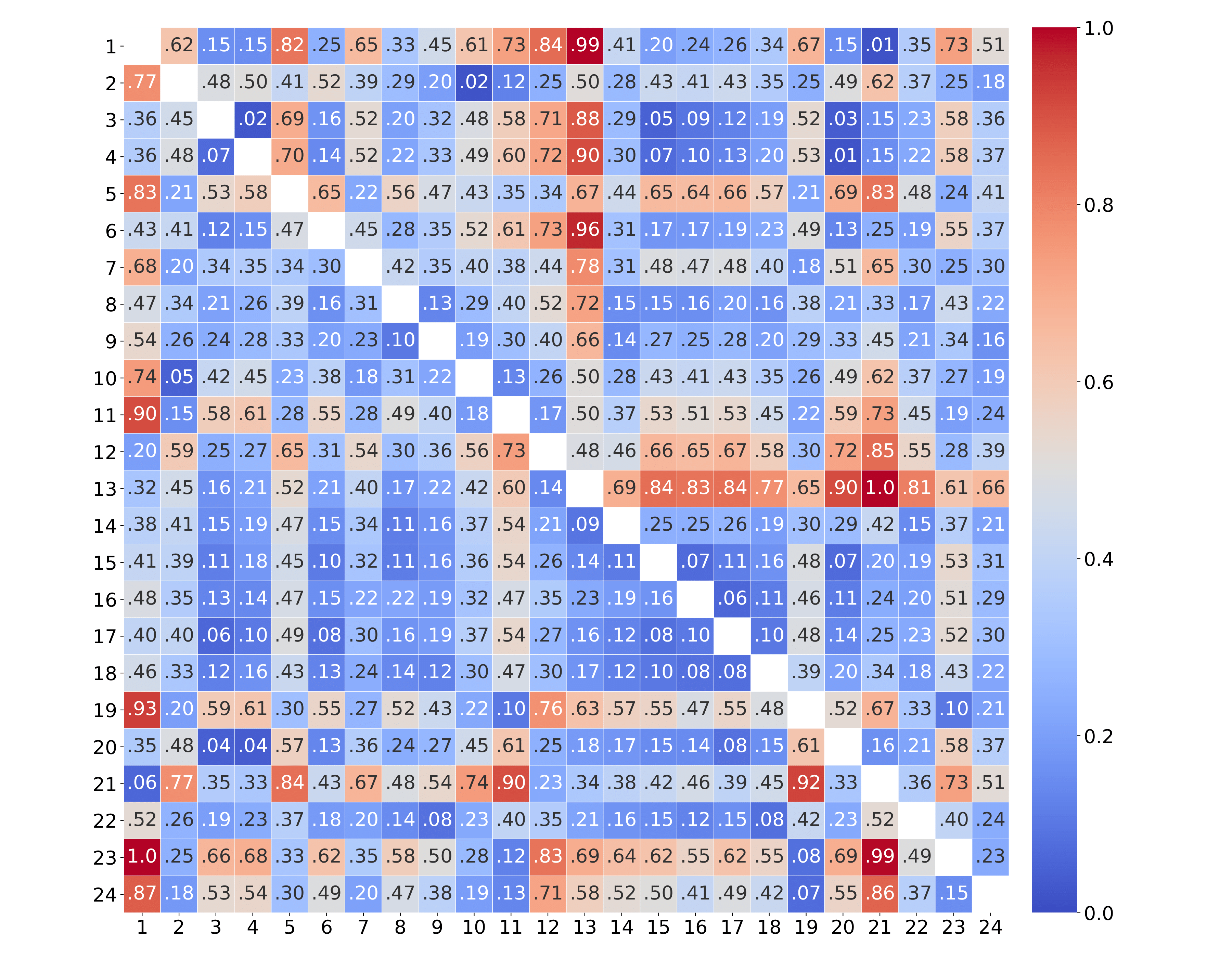}
\caption{ Pairwise distance matrices for all functions using data from the Standard CMA-ES. Euclidean distance is computed and scaled though min-max normalization. The lower triangle of the figure shows the pairwise distances for $L=100$, the upper triangle shows $L=500$.}
\label{fig:distFunAll}%
\end{figure}

The distance among all test problems is only computed from features of the standard CMA-ES in line with the classification setup (see section~\ref{sec:classification}). In Fig.~\ref{fig:distFunAll} a pairwise distance matrix is showing this result. We also drawn the result of a hierarchical clustering in Fig.~\ref{fig:tanglegramvariants} (on the right). From those two figures, we can see that the most closely related functions seem to be the triplet of functions $f_3$, $f_4$, and $f_{20}$, which is also very close to $f_{15}, f_{16}, f_{17}$, and $ f_{18}$ and hence forms a bigger cluster together. Besides, problem $f_1$ and $f_{21}$ are also decently similar, opening an interesting question on interpreting it since $f_{1}$ is generally considered simple while $f_{21}$ is conceivably difficult. We hypothesize that this is because the local topology of $f_{21}$ resembles the global topology of $f_1$. 

Another intriguing observation is that problem $f_{13}$ is noticeably different from the other problems for $L=500$ while such a difference does not exists for the $L=100$ case. This indicates that the dynamics of the standard CMA-ES when optimizing $f_{13}$ are similar to the other functions in the first $100$ generations of the optimization procedure and it that differentiates substantially from the others in the next $400$ generations. This entails that there might be huge changes to the landscape of $f_{13}$ between generations $100$ and $500$, attributing to this huge change in the CMA-ES's behavior. Interestingly, this connotation matches the design rationale behind $f_{13}$~\cite{bbobdef}: \textit{``Approaching the ridge is initially effective, but becomes ineffective close to the ridge... The necessary change in `search behavior' close to the ridge... ''}, hinting that a solver needs to change its search behaviour as it approaches the ridge defined in the function landscape. 

\section{Classifying CMA-ES variants and problems} \label{sec:classification}
Taking the 32 selected features, we constructed Extremely Randomized Trees Classifiers (ERTCs) to discriminate CMA-ES variances and the test problems for both cases when $L=100$ and $L=500$. This allows for investigating the impact of the length of raw time series, in terms of the classification power. Also, the resulting models are assessed using cross validation (CV) strategy (please see below for the detail). In addition to model accuracy we measure F1-score. Specifically, we calculate, for each fold, the precision ($p$) and recall $r$ with respect to each label  and take the harmonic mean, i.e. $2 \frac{p \times r}{p+r}$.

\paragraph{Classifying CMA-ES variants}
Here, we consider a multi-class classification task to distinguish CMA-ES variants on
\begin{itemize}
    \item each test problem separately, in which we have $|\mathcal{A}|\times N = 1200$ data points in each case, and 
    \item on the whole data set from all 24 test problems. 
\end{itemize}
For the former scenario, we perform a K-fold CV (with K = 24) and summarize the resulting classification accuracy in table~\ref{tab:resperfunc}, in which we observe an average accuracy of $.803\pm.087$ for $L=100$ and $.937\pm.069$ for $L=500$, indicating that a longer time series improves the classification performance. In detail, the lowest accuracy is observed on problem $f_{23}$ for both $L=100$ and $L=500$ while the highest one is achieved on $f_{21}$ for $L=100$ and $f_{6}$ for $L = 500$.

\begin{figure}
    \includegraphics[width=0.5\textwidth, trim={18mm 13mm 16mm 3cm},clip]{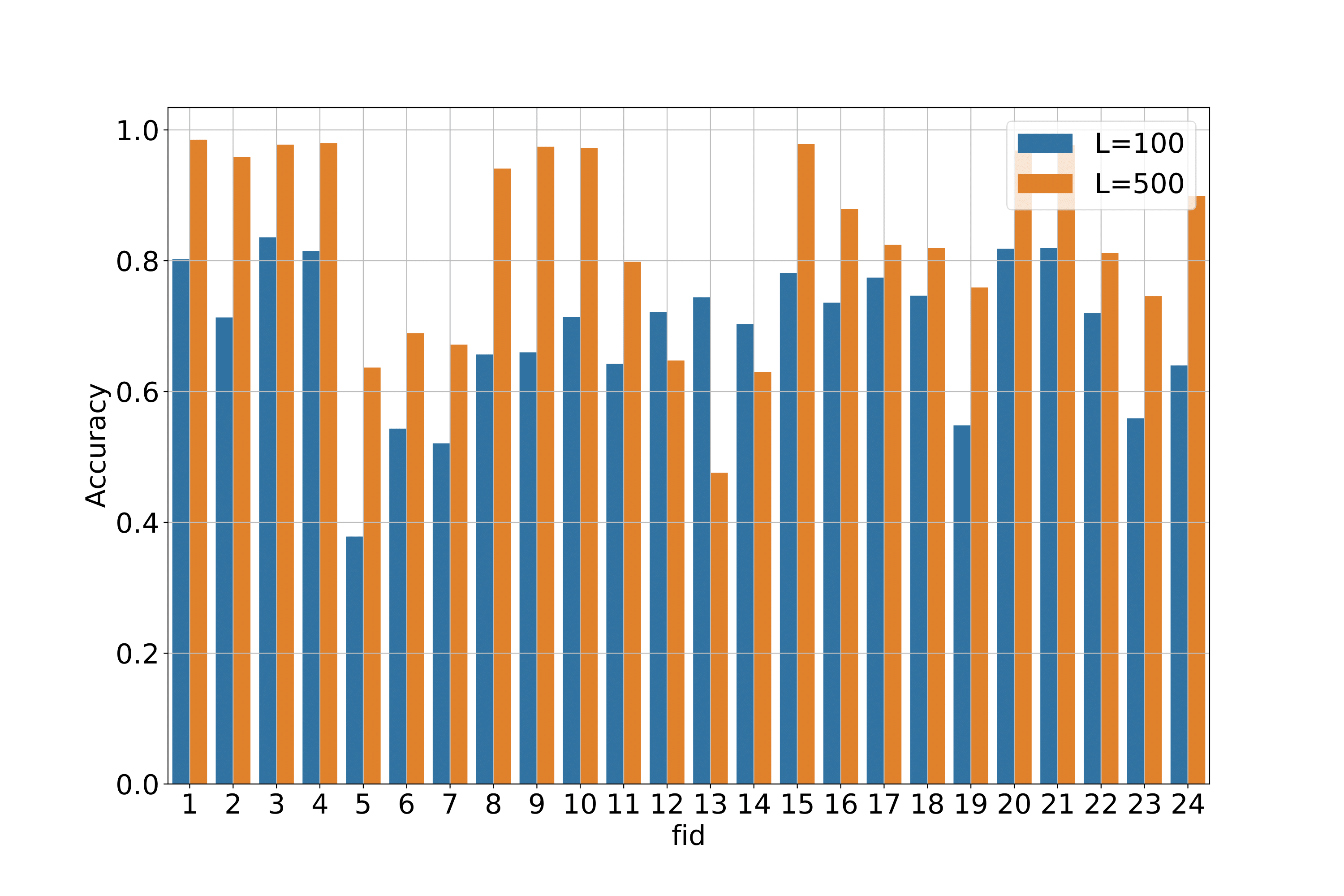}
    \caption{On classifying the CMA-ES variants over all test problems, Leave-One-Group-Out (LOGO) results show the testing accuracy on each single test problem, from a ERTC model trained on the remaining 23 problems. The $L=500$ case is consistently better than $L=100$ while those two cases exhibit quite similar trends over test problems.}
    \label{fig:accuracyperfunc}
\end{figure}

For the latter scenario, we merged data points from all $24$ test problems and conducted two different CV strategies: a standard randomized K-fold CV with $K=24$ and a \emph{Leave-One-Group-Out} (LOGO) CV, where we hold data points from a single problem for validation and train the model on the remaining $23$ problems. Note that the LOGO procedure iterates over each test problem. We show the resulting accuracy and F1 score in table~\ref{tab:stackedvarres}, in which we see that compared to both performance values in the K-fold case, the ones from LOGO CV drop slightly. This matches our expectation that the model would deteriorate if we exclude entirely the information of one test problem during the training and manage to generalize to it. We also notice that the variability of both performance values becomes much larger in the LOGO case, which is caused by substantial performance differences between each test problems. We support this statement with accuracy values in Fig.~\ref{fig:accuracyperfunc}, where the model poorly performs on $f_5$, $f_6$, and $f_7$ while its accuracy exceeds 0.95 on problems $f_1$-$f_4$, $f_{15}$, $f_{21}$ and $f_{22}$. Moreover, it is interesting that $L=500$ does not always outperform $L=100$ in terms of classification accuracy, e.g., on $f_{12}, f_{13}$, and $f_{14}$, $L=100$ it is noticeably higher.

 \begin{table}
 \small
 \begin{tabular}{l|ll||ll}
    \hline
     \textbf{FID} & \textbf{ Acc.} $L=100$ &  \textbf{Acc.} $L=500$ &  \textbf{F1} $L=100$ & \textbf{ F1} $L=500$\\ \hline 
     $f_{1}$  & .939 $\pm$ .032 & .991 $\pm$ .014 & .939 $\pm$ .034 & .991 $\pm$ .014 \\
     $f_{2}$  & .742 $\pm$ .065 & .980 $\pm$ .022  & .741 $\pm$ .071 & .980 $\pm$ .022  \\
     $f_{3}$  & .847 $\pm$ .057 & .983 $\pm$ .018 & .843 $\pm$ .059 & .984 $\pm$ .018 \\
     $f_{4}$  & .821 $\pm$ .050  & .983 $\pm$ .018 & .819 $\pm$ .050  & .983 $\pm$ .018 \\
     $f_{5}$  & .793 $\pm$ .064 & .822 $\pm$ .070  & .789 $\pm$ .072 & .825 $\pm$ .075 \\
     $f_{6}$  & .844 $\pm$ .053 & \textbf{.995 $\pm$ .009} & .842 $\pm$ .054 & \textbf{.995 $\pm$ .009} \\
     $f_{7}$  & .829 $\pm$ .050  & .878 $\pm$ .054 & .829 $\pm$ .055 & .877 $\pm$ .058 \\
     $f_{8}$  & .771 $\pm$ .062 & .971 $\pm$ .024 & .764 $\pm$ .062 & .971 $\pm$ .024 \\
     $f_{9}$  & .788 $\pm$ .066 & .982 $\pm$ .022 & .782 $\pm$ .067 & .982 $\pm$ .024 \\
     $f_{10}$ & .771 $\pm$ .042 & .986 $\pm$ .015 & .765 $\pm$ .046 & .986 $\pm$ .015 \\
     $f_{11}$ & .796 $\pm$ .076 & .992 $\pm$ .013 & .795 $\pm$ .080  & .992 $\pm$ .013 \\
     $f_{12}$ & .877 $\pm$ .039 & .945 $\pm$ .020  & .877 $\pm$ .040  & .946 $\pm$ .019 \\
     $f_{13}$ & .828 $\pm$ .051 & .881 $\pm$ .043 & .826 $\pm$ .058 & .882 $\pm$ .041 \\
     $f_{14}$ & .870 $\pm$ .044  & .935 $\pm$ .037 & .868 $\pm$ .043 & .936 $\pm$ .036 \\
     $f_{15}$ & .809 $\pm$ .051 & .975 $\pm$ .027 & .812 $\pm$ .053 & .976 $\pm$ .027 \\
     $f_{16}$ & .800 $\pm$ .051   & .944 $\pm$ .026 & .804 $\pm$ .052 & .944 $\pm$ .027 \\
     $f_{17}$ & .788 $\pm$ .057 & .878 $\pm$ .047 & .788 $\pm$ .056 & .877 $\pm$ .048 \\
     $f_{18}$ & .748 $\pm$ .057 & .850 $\pm$ .056  & .748 $\pm$ .061 & .846 $\pm$ .055 \\
     $f_{19}$ & .680 $\pm$ .052  & .845 $\pm$ .052 & .681 $\pm$ .057 & .842 $\pm$ .051 \\
     $f_{20}$ & .834 $\pm$ .049 & .978 $\pm$ .016 & .834 $\pm$ .051 & .979 $\pm$ .016 \\
     $f_{21}$ & \textbf{.958 $\pm$ .033} & .992 $\pm$ .021 & \textbf{.958 $\pm$ .033} & .993 $\pm$ .021 \\
     $f_{22}$ & .767 $\pm$ .049 & .939 $\pm$ .034 & .767 $\pm$ .052 & .941 $\pm$ .033 \\
     $f_{23}$ & \underline{.677 $\pm$ .072} &\underline{.810 $\pm$ .063}  &\underline{.670 $\pm$ .077}  &\underline{.807 $\pm$ .066} \\
     $f_{24}$ & .705 $\pm$ .066 & .952 $\pm$ .035 & .704 $\pm$ .070  & .953 $\pm$ .037 \\
    \hline   
    \textbf{Avg.} & .803 $\pm$ .067 & .937 $\pm$ .058 & .802 $\pm$ .068 & .937 $\pm$ .059 \\ 
    \hline \hline
\end{tabular}
    \caption{For classifying CMA-ES variants on each test problem separately, we list the accuracy and F1 scores resulted from a $24$-fold CV procedure with the highest measured values marked in boldface and the lowest ones underlined.}
    \label{tab:resperfunc}
\end{table}
\begin{table}[ht]
 \small
 \begin{tabular}{l|l|l|l|l} 
    \hline
    \textbf{CV} & \textbf{ Acc.} $L=100$ &  \textbf{Acc.} $L=500$ &  \textbf{F1} $L=100$ & \textbf{ F1} $L=500$\\ \hline 
    K-Fold  & .769 $\pm$ .013   &  .923 $\pm$ .005 &  .766 $\pm$  .014 &  .923 $\pm$ .005  \\ \hline
    LOGO    & .691 $\pm$ .114   & .833 $\pm$ .148  &  .682 $\pm$ .115  & .822 $\pm$ .120 \\ \hline
    \bottomrule 
 \end{tabular}
 \caption{Comparison of two classification approaches for CMA-ES variants: on each test problem individually (see table~\ref{tab:resperfunc}) and on all problems jointly (see Fig.~\ref{fig:accuracyperfunc}).}
 \label{tab:stackedvarres}
\end{table}

\paragraph{Classifying test problems}
Furthermore, we delve into the question that whether the time series features are capable of discriminating test problems, which, if so, could serve as an alternative to the well-known Explorative Landscape Analysis (ELA) features~\cite{kerschke2019comprehensive} for identifying problem classes. For this purpose, we took only the time series feature data from the \emph{Standard} CMA-ES to distinguish test problems using the ERTC method\footnote{This choice is made based on our observation that all CMA-ES variants give rise to very similar classification results. Hence, it is more natural to take the baseline variant.}. As with the previous task, we perform classification for both cases when $L=100$ and $L=500$, and use $24$-fold cross validation.

\begin{figure*}
    \includegraphics[width=\textwidth]{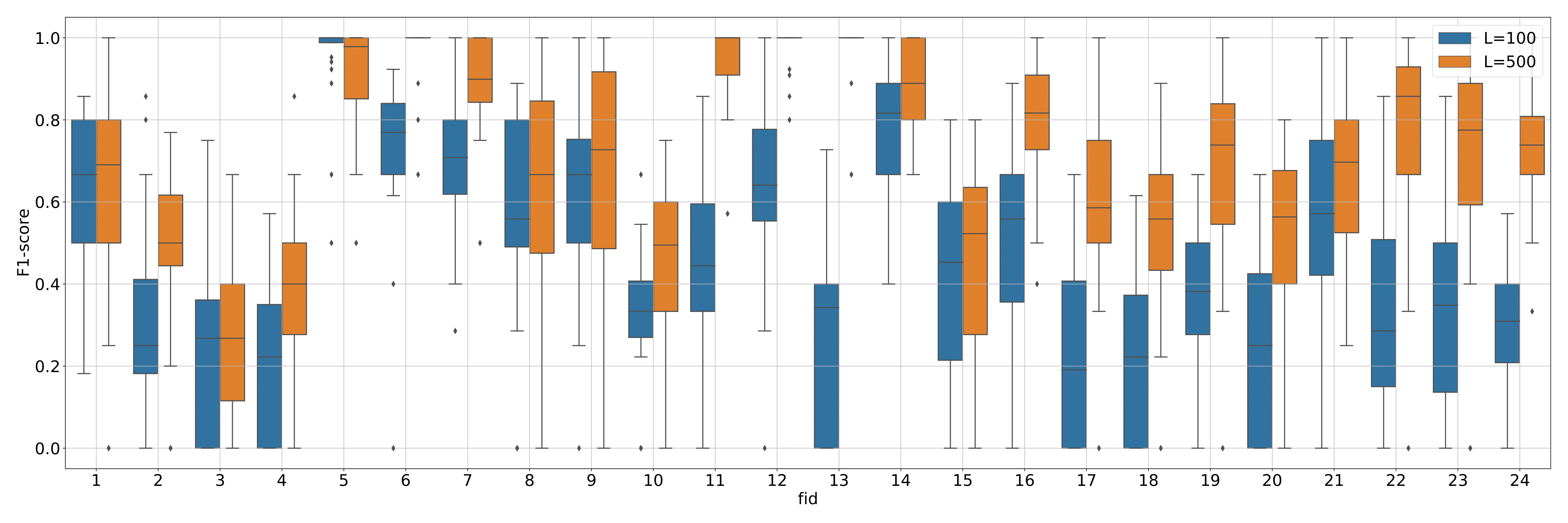}
    \caption{On classifying the test problems, we illustrate the resulting F1-score per each classification label, where the box plot indicates the variability of this score across a 24-fold cross validation. We also contrast the $L=100$ case with the $L=500$ one.}
    \label{fig:boxplotperfun}
\end{figure*}

This experiment, however, only achieves an accuracy of $.497\pm.053$ for the $L=100$ while it manages to reach $.716\pm.048$ for $L=500$. In contrast to the study on classifying algorithms, we encounter a substantial performance drop. In figure~\ref{fig:boxplotperfun}, a box plot is shown for this experiment, which shows the F1-score per classification label obtained for each cross validation fold. Again, we observe a large variance between the labels. Interestingly, we observe a pattern opposite of that of figure~\ref{fig:accuracyperfunc}. Now, $f_5$ is one of the easiest functions to predict, while when using data from $f_5$ for testing in the LOGO experiment, rather poor classification performance was observed. This is in line with our earlier observation from the LOGO experiment; when a classifier is not able to learn function specific patterns in the data, it shows lower performance in classifying variants. The fact that $f_5$ is classified correctly more often than most other functions in this experiment, means that there is a pattern in the data of $f_5$ which distinguishes it from the data of other functions. This then also means, that when using data of $f_5$ for testing in the LOGO experiment, the classifier has a harder time giving a correct prediction, because it is now working on data which is very dissimilar from the data is was trained on. The reverse can also be observed in $f_3$. Because there are other functions in the data which express similar patterns, such as $f_4$, the classifier is not able to clearly distinguish between them. However, when predicting variants (figure~\ref{fig:accuracyperfunc}), the classifier is able to learn these function specific patterns during training from $f_4$, which in turn allows it to more confidently predict variants when using data taken from $f_3$ for testing.

In addition to classify each test problem, we perform a classification task on problem groups, where the problems are partitioned according to the default groups in COCO/BBOB, or to the clustering results in~\cite{mersmanBBOB} (referred as Mersman's grouping hereafter), which is obtained on the expected running time (ERT) values of $51$ different optimization algorithms on $5D$. The latter grouping might be particularly suitable for this study, since it is obtained based on the consensus rankings of different optimization algorithms, which reflects, to some extent, the algorithmic behaviour and hence could be well related to the time series features extracted from dynamics of the algorithm. This conjecture is empirically supported by table~\ref{tab:funcgroup}, where the classification outcome of Mersman's grouping is much better than that of the default BBOB grouping for $L=100$ and $L=500$.
\begin{table}[!t]
\hspace*{-3mm}
\small
\begin{tabular}{l|l|l|l|l} 
\hline
\textbf{Groups} & \textbf{ Acc.} $L=100$ &  \textbf{Acc.} $L=500$ &  \textbf{F1} $L=100$ & \textbf{ F1} $L=500$\\ \hline 
BBOB        & .643 $\pm$ .054   & .758 $\pm$ .050  & .640 $\pm$ .058  & .750 $\pm$ .053 \\ \hline
Mersmann    & .774 $\pm$ .058   & .840 $\pm$ .114  & .756 $\pm$ .060  & .820 $\pm$  .124\\ \hline
\bottomrule 
\end{tabular}
\caption{Results on classifying function groups with a $24$-fold CV procedure for both $L=100$ and $L=500$ cases, in which the feature extracted from the standard CMA-ES is used.}
\label{tab:funcgroup}
\end{table}

\section{Predicting CMA-ES's performance} \label{sec:regression}
From figure~\ref{fig:distributionall} (right), we see clearly that feature vectors that are close to each other in terms of its two-dimensional embedding appear to be associated with similar numbers of targets reaches, which motivates us to predict the performance of each run using the corresponding features. For this goal, we adopted an Extremely Randomized Trees Regressor (ERTR, from scikit-learn~\cite{sklearn}) with its default parameter setting. The independent variables in this regression experiment consist of the $32$ selected time series features, which are taken as input to predict the number of targets $N_s$. Also, we took the coefficient of determination $R^2(y, \hat y) = 1 - \frac{\sum(y - \hat y)^2}{\sum(y - mean(y))^2}$ as the performance metric and applied a $24$-fold CV procedure as before, for both $L=100$ and $L=500$ cases. 

\begin{figure}[!ht]
    \centering
    \includegraphics[width=\linewidth,trim={.8cm 0 1cm 0},clip]{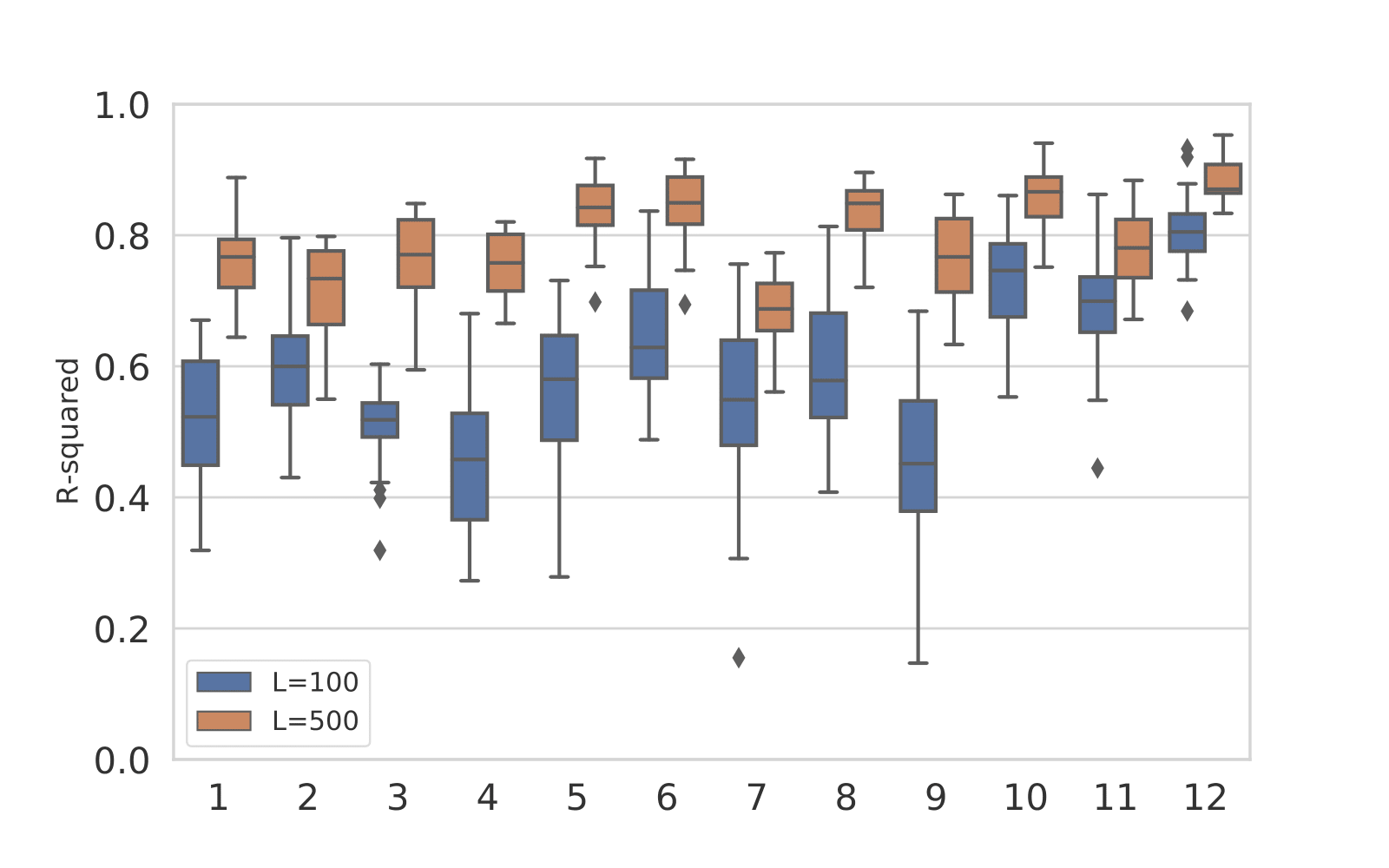}
    \caption{Regression, with a model trained on each CMA-ES variant separately, showing a box plot of the $R^2$ attained using K-fold cross validation, with K=24. The numeric index used here on the x-axis refers to the number specified in the list in section~\ref{sec:background}. Data for $L=100$ is shown in blue, $L=500$ is shown in orange.}
    \label{fig:regperfunc}%
\end{figure}

When training the ERTR model on all variants (on all problems), we observed an $R^2$ of $.557\pm.018$ for $L=100$, and $.778\pm .014$ for $L=500$. Alternatively, when training the model for each variants separately (see Fig.~\ref{fig:regperfunc}), we see the a huge improvement of the $R^2$ on the $L=500$ case for most variants, which is also considerably stable in contrast to the $L=100$ case. It is worth noting that we do not consider a scenario in which we took each problem separately, since the distribution of $N_s$ value varies drastically from function to function. This is because the CMA-ES shows a different performance on each benchmark function, signifying that data from a given function can not be used to train a regression model for another function, and vice versa. 

\section{Conclusions} \label{sec:conclusion}
In this work, we have used time series characterization methods to extract features from the dynamic strategy parameters of the CMA-ES variants during the optimization of 24 test problems from the Black-Box Optimization Benchmark (BBOB). After extracting the time series features on $5D$ with two different running length of each variant ($L=100$ and $L=500$ for 100 and 500 iterations, respectively), we conducted a proper feature selection to reduce the dimensionality of the resulting features. We built extremely randomized trees to classify the variants on a single and multiple problems, as well as to distinguish those test problems. Furthermore, we performed in-depth explorative data analysis on the selected features, where we show the hierarchical cluster discovered in CMA-ES variants and in test problems and contrast such clustering results obtained in both $L=100$ and $L=500$ cases. Moreover, we explored the potential of directly predicting the performance of CMA-ES variants using their corresponding features.

We observed that the time series feature are capable of discriminating CMA-ES variants with nearly perfect accuracy, if we train the model on a single problem only. We also managed to achieve a decent accuracy when classifying the function groups. When it comes to predict the performance of those variants, we saw an indication that the feature might be reasonably related to the performance, which is subject to future researches.

From the clustering analysis of test problems, we observed an intriguing pattern on problem $f_{13}$, which resembles many other problems for the $L=100$ case and becomes very distant from all other functions for $L=500$. This phenomenon suggests that there might be a drastic change in the landscape which causes the change of the behavior of the CMA-ES, which is reflected by the feature. This is also supported by the fact that the landscape of $f_{13}$ is very heterogeneous.

For the future research, we are planning to 1) apply this approach on more problems and algorithms; 2) to investigate the usefulness of the time series feature for the dynamic algorithm selection task; 3) to extend it by computing the features in a moving window; and 4) to dive more deeply into the data to investigate the relative contribution and values of each feature with the help of, for instance the Shapley value.

\bibliographystyle{ACM-Reference-Format}
\bibliography{main} 
\end{document}